\documentclass{article}
\usepackage{ICASSP2022,amsmath,graphicx}

\usepackage{booktabs}
\usepackage{multirow} 
\usepackage{amssymb} 
\usepackage{xcolor}
\usepackage{hyperref} 
\hypersetup{
    colorlinks,
    linkcolor={black},
    citecolor={black},
    urlcolor={black}
}
\usepackage{float}
\usepackage{dirtytalk}
\usepackage{pifont}
\newcommand{\cmark}{\ding{51}}%
\newcommand{\xmark}{\ding{55}}%
\usepackage[backend=bibtex,citestyle=numeric-comp,bibstyle=ieee,sorting=none,defernumbers=true,giveninits=true,doi=false,isbn=false,url=false,eprint=false,minbibnames=3,maxbibnames=3]{biblatex}
\usepackage{xcolor}

\usepackage[normalem]{ulem}
\usepackage{tikz}
\newcommand*\circled[1]{\tikz[baseline=(char.base)]{
            \node[shape=circle,draw,inner sep=0.3pt] (char) {#1};}}

\usepackage{tabularx}
\usepackage{ragged2e}
\usepackage{bm}
\usepackage{boldline}

\newcolumntype{C}{>{\Centering\arraybackslash}X}

\newcolumntype{u}{>{\raggedright\hsize=.7\hsize}X}
\newcolumntype{t}{>{\Centering\hsize=.6\hsize}X}
\newcolumntype{s}{>{\Centering\hsize=.5\hsize}X}
\newcolumntype{j}{>{\Centering\hsize=.4\hsize}X}
\newcolumntype{k}{>{\Centering\hsize=.3\hsize}X}
\newcolumntype{y}{>{\Centering\hsize=.2\hsize}X}
\newcolumntype{g}{>{\Centering\hsize=.16\hsize}X}
\newcolumntype{z}{>{\Centering\hsize=.1\hsize}X}
\newcolumntype{e}{>{\raggedright\hsize=.35\hsize}X}
\newcolumntype{v}{>{\raggedright\hsize=.55\hsize}X}
\newcolumntype{f}{>{\raggedright\hsize=.32\hsize}X}
\newcolumntype{q}{>{\raggedright\hsize=.68\hsize}X}

\usepackage{subcaption}

\title{Training Strategies For Improved Lip-reading}

\name{
    Pingchuan Ma$^{1*}$,
    Yujiang Wang$^{1*\dagger}$,
    Stavros Petridis$^{1,2}$,
    Jie Shen$^{1,2}$,
    Maja Pantic$^{1,2}$\thanks{$^{*}$ Equal contribution.}
    \thanks{$^{\dagger}$ Corresponding author (e-mail: \href{mailto:yujiang.wang14@imperial.ac.uk}{yujiang.wang14@imperial.ac.uk}).}
    \thanks{All training, testing and ablation studies have been conducted at Imperial College London.}
    \thanks{Code and trained models are available at: \url{https://sites.google.com/view/audiovisual-speech-recognition} }
}
\address{
$^1$ Imperial College London, UK \\
$^2$ Meta AI, UK
}%
\begin{document}
%
\maketitle
\begin{abstract}
Several training strategies and temporal models have been recently proposed for isolated word lip-reading in a series of independent works. However, the potential of combining the best strategies and investigating the impact of each of them has not been explored. In this paper, we systematically investigate the performance of state-of-the-art data augmentation approaches, temporal models and other training strategies, like self-distillation and using word boundary indicators. Our results show that Time Masking (TM) is the most important augmentation followed by mixup and Densely-Connected Temporal Convolutional Networks (DC-TCN) are the best temporal model for lip-reading of isolated words. Using self-distillation and word boundary indicators is also beneficial but to a lesser extent. A combination of all the above methods results in a classification accuracy of 93.4\%, which is an absolute improvement of 4.6\% over the current state-of-the-art performance on the LRW dataset. The performance can be further improved to 94.1\% by pre-training on additional datasets. An error analysis of the various training strategies reveals that the performance improves by increasing the classification accuracy of hard-to-recognise words.

\end{abstract}
\vspace{-0.3cm}
\section{Introduction}
\label{sec:intro}
\vspace{-0.2cm}
Lip-reading of isolated words has received a lot of attention recently thanks to the availability of large publicly available datasets like LRW~\cite{chung16}. The majority of works follow the same lip-reading pipeline consisting of a visual encoder, followed by a temporal model and a softmax classification layer. The visual encoder proposed by~\cite{stafylakis2017combining} has been widely adopted in most works,  hence, most recent efforts aim at improving the temporal model or the training strategy. Bidirectional Gated Recurrent Units (BGRUs) and Multi-Scale Temporal Convolutional Networks (MS-TCNs) have been the most popular temporal models in the literature and conflicting conclusions about their performance have been reported. For example, MS-TCNs outperformed BGRUs in~\cite{martinez2020lipreading} but not in~\cite{DBLP:journals/corr/abs-2011-07557}. Similarly, different data augmentations have been presented like mixup~\cite{DBLP:journals/corr/abs-2011-07557, ma2020towards}, variable length augmentation~\cite{martinez2020lipreading} and cutout~\cite{zhang2020can}. Other improvements which have been proposed in the literature include the addition of word boundary indicators~\cite{stafylakis18}, which  define the start and end frame of a word in a video, and self distillation~\cite{ma2020towards} which results in a series of networks with the same architecture trained via distillation. All these improvements have been proposed separately in the literature and a study combining all of them and investigating the impact of each of them is missing.

In this work, we present a model trained with some of the most promising recent ideas and evaluate the contribution of each of them via an ablation study. This is a useful study since we can quantify the effect of each method when combined with other augmentation methods or temporal models. We also provide an error analysis demonstrating how each method improves the lip-reading accuracy. To the best of our knowledge, the only similar study that exists is~\cite{DBLP:journals/corr/abs-2011-07557} but despite using some of the latest methods it was only able to match the current state-of-the-art performance.     

Our results demonstrate that: 1) We can achieve a new state-of-the-art performance on the LRW dataset by combining all the latest data augmentation methods, using the recently proposed DC-TCN, word boundary indicators and self-distillation. The accuracy achieved is 92.8\% for a single model and 93.4\% for an ensemble. The performance can be slightly improved to 93.5\% and 94.1\%, respectively, by pre-training on additional datasets. 2) Time Masking is the most effective augmentation method followed by mixup. The use of DC-TCN significantly outperforms the MS-TCN which in turn outperforms the BGRU model. The use of word boundary indicators and self-distillation is also beneficial with the former resulting in greater improvement. 3) The error analysis suggests that all these methods improve performance by significantly increasing the classification accuracy of difficult words.

\begin{figure*}[!t]
    \centering
    \begin{subfigure}{.31\linewidth}
        \centering
        \includegraphics[height=200pt]{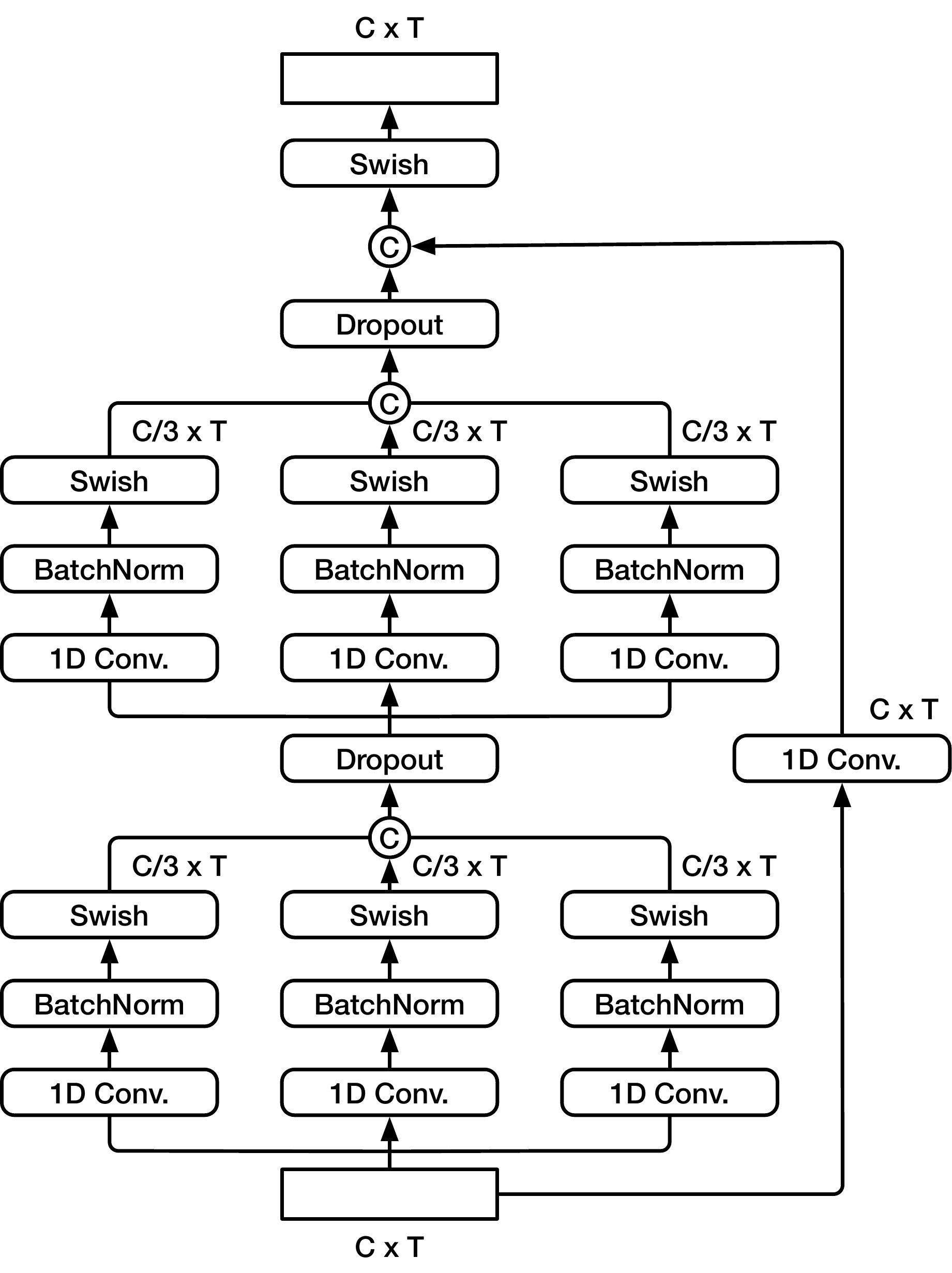}
        \caption{}
        \label{fig:mstcn}  
    \end{subfigure}
    \qquad
    \begin{subfigure}{.31\linewidth}
        \centering
        \includegraphics[height=200pt]{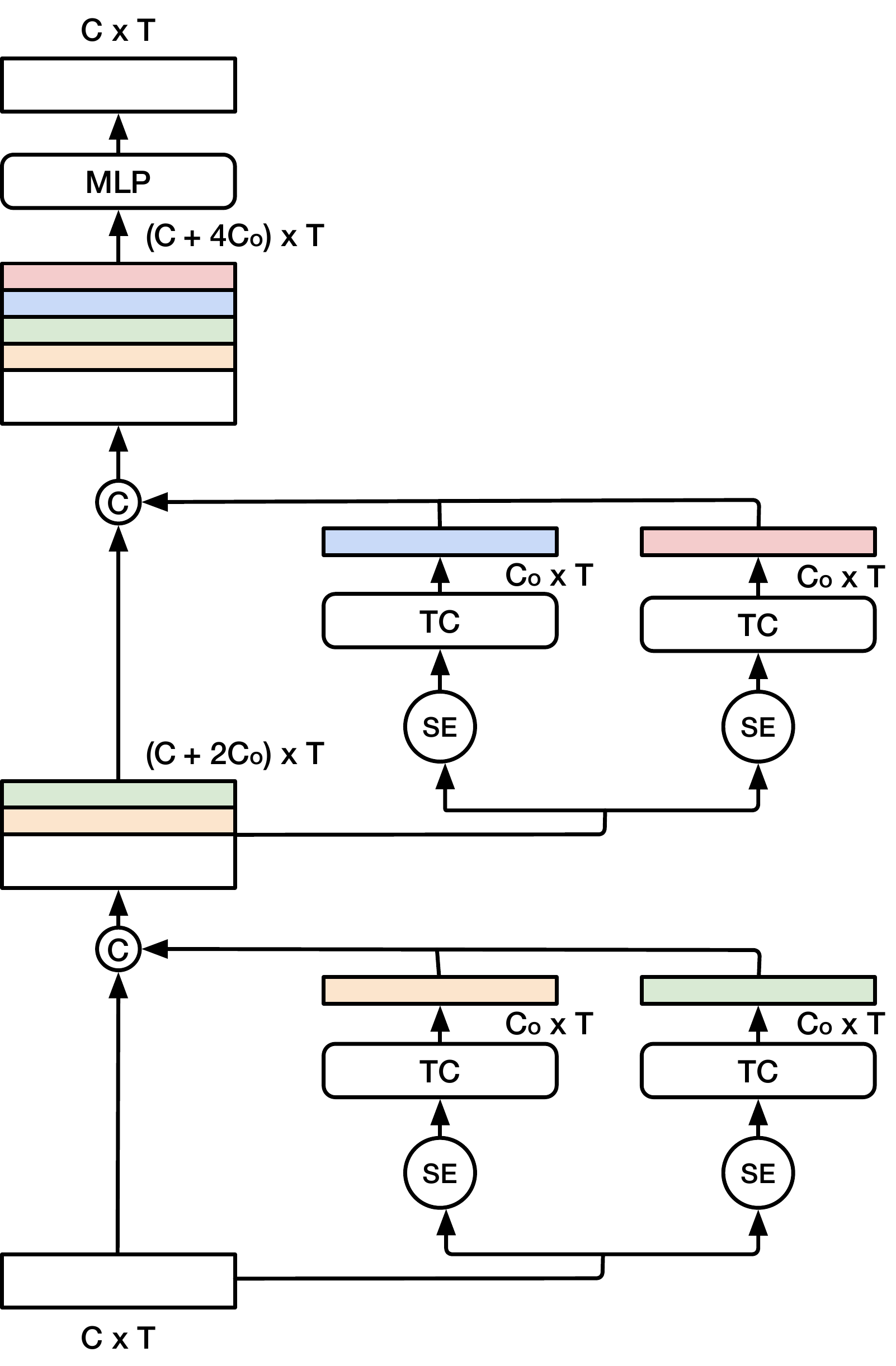}
        \caption{}
        \label{fig:dc_tcn}  
    \end{subfigure}
    \quad
    \begin{subfigure}{.31\linewidth}
        \centering
        \includegraphics[height=200pt]{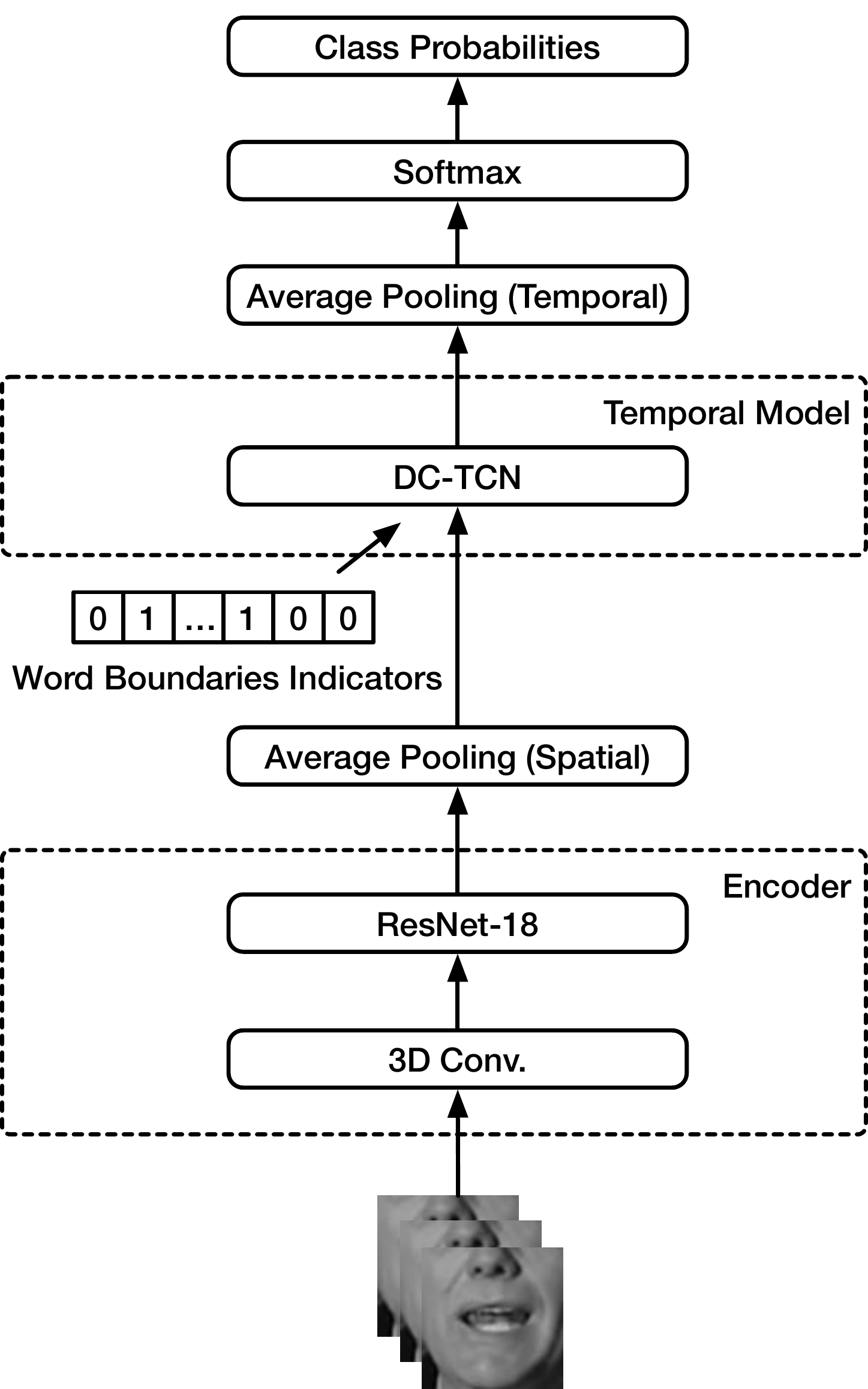}
        \caption{}
        \label{fig:c}  
    \end{subfigure}
    \caption[architecture]{(a): MS-TCN architecture. \protect\say{$C$} and \protect\say{$T$} refer to the channel number and sequence length, respectively. (b): DC-TCN architecture. \circled{{\footnotesize SE}} and \circled{{\footnotesize C}} denote the the operations of Squeeze-and-Excitation (SE) \cite{hu2018squeeze} and channel-wise concatenation, respectively. \protect\say{$TC$} represents a Temporal Convolutional block, while the growth rate is denoted as \protect\say{$C_o$}.
(c): Lip-reading model with a modified ResNet-18 as encoder and DC-TCN as a temporal model. The word boundary indicators are concatenated with the output features of the encoder.}
\label{fig:architecture}
\vspace{-0.5cm}
\end{figure*}
\vspace{-0.3cm}
\section{Training Strategies }
\noindent\textbf{Architecture}\quad
The first building block of the model (Fig.~\ref{fig:c}) is the most commonly used mouth Region-Of-Interest (ROI) encoder consisting of a 3D convolutional layer, which takes as input 5 consecutive frames, followed by a 2D ResNet-18~\cite{stafylakis2017combining}. The frame-wise features from the output of the encoder are then fed to a temporal model to capture the temporal dependencies. This is followed by a softmax layer which outputs the class probabilities over the words to be classified.

In this work, we investigate the impact of three different temporal models for the recognition of isolated words, BGRUs~\cite{petridis2018end}, MS-TCNs~\cite{martinez2020lipreading} and DC-TCNs~\cite{ma2021lip}.  TCNs consist of a stack of temporal convolutional (TC) blocks , where each block consists of a few layers of dilated convolutions with kernel size $k$. A MS-TCN (Fig. \ref{fig:mstcn}) extends the vanilla TCN by adding multiple branches each with different kernel sizes, and the features from the output of each branch are concatenated to mix information at several temporal scales. A DC-TCN (Fig. \ref{fig:dc_tcn}) extends the vanilla TCN by adding dense connection at each TC block and using a Squeeze-and-Excitation (SE) attention mechanism.

\noindent\textbf{Data Augmentation}\label{ssec:DataAugm}\quad
\textit{Random Cropping:} We randomly crop an 88 $\times$ 88 patch from the mouth ROI during training. At test time, we simply crop the central patch. This is a commonly used augmentation method that has been used successfully in several lip-reading works~\cite{martinez2020lipreading, petridis2018end}.    
\textit{Flipping:} We randomly flip all the frames in a video with a probability of 0.5. This augmentation is commonly used in combination with random cropping~\cite{martinez2020lipreading, petridis2018end}.  
\textit{Mixup:} We create new augmented training examples by linearly combining two input video sequences  and their corresponding targets. 
We set the linear combination weight $\lambda$ to be $0.4$ similarly to~\cite{ma2020towards}.
\textit{Time Masking:} We mask $N$ consecutive frames for each training sequence where $N$ is sampled between $0$ and $N_{\text{max}}$ using a uniform distribution. Each masked frame is replaced with the mean frame of the sequence it belongs to. This augmentation is based on SpecAugment~\cite{DBLP:conf/interspeech/ParkCZCZCL19}, which has been proposed for ASR applications, and aims at making the model more robust to small segments with missing frames.  

\begin{table*}[t]
\centering
\small
\begin{tabularx}{0.92\linewidth}{k z z z z g g y k g}
\toprule
\multirow{2}{*}[-0.0em]{\textbf{\shortstack{Temporal Model}}} &
\multicolumn{4}{c}{\textbf{Data Augmentation}} &
\multirow{2}{*}[-0.0em]{\textbf{\shortstack{Word \\Boundary}}} &
\multicolumn{3}{c}{\textbf{Pre-training Strategies}} &
\multirow{2}{*}[-0.0em]{\shortstack{\textbf{Top-1} \\ \textbf{Acc.} (\%)} } \\ 
& Crop & Flip & Mixup & TM & & Scratch & LiRA(LRS3) & LRS2\&3+AVS & \\ 
\cmidrule(l{4pt}r{4pt}){1-1}
\cmidrule(l{4pt}r{4pt}){2-5} 
\cmidrule(l{4pt}r{0pt}){6-6}
\cmidrule(l{4pt}r{0pt}){7-9}
\cmidrule(l{4pt}r{4pt}){10-10} 
\multirow{8}{*}[-0.4em]{DC-TCN \cite{ma2021lip}} & \cmark & \cmark & \cmark & \cmark & \cmark  & - & - &\cmark & \textbf{92.9} \\
& \cmark & \cmark & \cmark & \cmark & \cmark & - & \cmark & - & 92.3 \\
& \cmark & \cmark & \cmark & \cmark & \cmark & \cmark & - &- & 92.1 \\
& - & \cmark & \cmark & \cmark & \cmark & \cmark & - &- & 91.8 \\
& \cmark & - & \cmark & \cmark & \cmark & \cmark & - &- & 91.7 \\
& \cmark & \cmark & - & \cmark & \cmark & \cmark & - &- & 91.0 \\
& \cmark & \cmark & \cmark & - & \cmark & \cmark& - &- & 89.7 \\
& \cmark & \cmark & \cmark & \cmark & - & \cmark & - &- & 90.4 \\ 
\cmidrule(l{4pt}r{4pt}){1-1}
\cmidrule(l{4pt}r{4pt}){10-10} 
MS-TCN \cite{martinez2020lipreading} &  \cmark &  \cmark &  \cmark & \cmark & \cmark & \cmark & - &- & 90.0 \\ 
\cmidrule(l{4pt}r{4pt}){1-1}
\cmidrule(l{4pt}r{4pt}){10-10} 
BGRU~\cite{petridis2018end} & \cmark & \cmark & \cmark & \cmark & \cmark & \cmark & - &- & 89.7  \\
\bottomrule
\end{tabularx}
\caption{Ablation studies of three  temporal models on LRW dataset. Starting from the best-performing DC-TCN model, we remove each data augmentation and the word boundaries indicators to examine their effectiveness. Then we replace the DC-TCN with MS-TCN and BGRU.  \protect\say{Scratch} denotes a model trained from scratch without using external data. 
\protect\say{LiRA(LRS3)}
indicates a self-supervised pre-trained model  using LiRA~\cite{DBLP:journals/corr/abs-2106-09171} on the LRS3 dataset, and \protect\say{LRS2\&3+AVS} indicates a fully supervised pre-trained model on LRS2, LRS3 and AVSpeech.}

\label{tab:distentangling experiments on lrw} 
\end{table*}
\noindent\textbf{Word Boundary Indicator}\quad
Following~\cite{DBLP:journals/corr/abs-2011-07557, stafylakis18}, we add word boundary indicators as extra input to the temporal model. The indicators are basically  binary vectors with the same length as the number of frames in the input video. All vector entries which correspond to frames where the target word is present are set to 1 and the rest are set to 0. The vector for the word boundary indicators is concatenated with the frame-wise visual features from the encoder and the new vector is fed into the temporal model.

\noindent\textbf{Self-Distillation}\quad
Self-distillation~\cite{furlanello2018born} is based on the idea of training a series of models with the same architecture using distillation and has been recently applied to lip-reading~\cite{ma2020towards}. Specifically, we first train a network that acts as a teacher for training a student model with the same architecture.
The student network becomes the teacher network in the next iteration and we keep training models until no improvement is observed. The insight behind this is that the teacher network provides extra supervisory signal with inter-class similarity information.
The overall loss $\mathcal{L}$ to be optimized is the weighted combination of Cross-Entropy loss $\mathcal{L}_{{\rm CE}}$ for hard targets and Kullback-Leibler (KL) divergence loss $\mathcal{L}_{{\rm KD}}$ for soft targets.
\begin{equation}
\mathcal{L} = \mathcal{L}_{{\rm CE}}(y, \delta(z_s;\theta_s)) + \alpha \mathcal{L}_{{\rm KD}}(\delta(z_s;\theta_s), \delta(z_t;\theta_t))
\label{eq:KD_loss}
\end{equation}
where $z_s$ and $z_t$ represent the embedded representations from student and teacher networks, respectively, $\theta_s$ and $\theta_t$ denote learnable parameters of student and teacher models,  $y$ is the target label, $\delta(\cdot)$ stands for the softmax function, and $\alpha$ is the balancing weight between the two terms.

\section{Experimental Setup}

\noindent\textbf{Databases}\quad
In our experiments, we employ LRW~\cite{chung16}, which is the largest publicly available dataset for lip-reading of isolated words. The dataset is collected in a form of short clips from more than 1\,000 speakers in BBC programs and contains 500 isolated words.
Each clip has a duration of 29 frames (1.16 seconds). The isolated word is centred within the clip. The dataset is composed of 488\,766, 25\,000, and 25\,000 short clips in the training, validation and test sets.

\noindent\textbf{Pre-Processing}\quad
We used the RetinaFace~\cite{DBLP:journals/corr/abs-1905-00641} tracker to detect the faces and the Face Alignment Network (FAN)~\cite{bulat2017far} for landmark detection. The size and rotation differences are removed through registering the faces to the mean face in the training set. A bounding box of 96 $\times$ 96 is used to crop the mouth ROIs. Each frame is normalised by subtracting the mean and dividing by the standard deviation of the training set.

\noindent\textbf{Training Details}\quad
The model is trained for 80 epochs with a mini-batch size of 32. 
We use the AdamW optimizer~\cite{loshchilov2017decoupled} with an initial learning rate of 3e-4. The learning rate is decayed using a cosine annealing strategy without a warm-up phase. We also use the variable-length augmentation~\cite{martinez2020lipreading} for all experiments. The value for $N_{\text{max}}$ used in time-masking (see section \ref{ssec:DataAugm}) is set to 15 frames (0.6 seconds) and was optimised in the LRW validation set. 

\noindent\textbf{Temporal Models}\quad
\textit{MS-TCN:} We adopt the same MS-TCN architecture as in~\cite{martinez2020lipreading}, that is, each block consists of 3 branches with 3/5/7 kernel sizes, respectively, and we stack 4 such blocks to formulate the MS-TCN network.
\textit{DC-TCN:}
The DC-TCN used in this paper generally follows the structures in~\cite{ma2021lip}. In particular, we opt for the Partially Dense (PD) architecture in each TC block, while each block consists of 9 densely-connected temporal convolutions with kernel sizes of \{3, 5, 7\} and dilation rates of \{1, 2, 5\}.  
\textit{BGRU:} A four-layer BGRU with a dropout rate of 0.2 is used with 1024 hidden units.

\noindent\textbf{Initialisation}\quad
To investigate the impact of initialisation we consider three cases: 1) we train the model from scratch using only the LRW training set, 2) we pre-train the encoder from Fig. \ref{fig:architecture} on the LRS3 dataset~\cite{Afouras18d} using the LiRA~\cite{DBLP:journals/corr/abs-2106-09171} self-supervised approach and fine-tune it on the LRW training set. 3) we pre-train the encoder on LRS2~\cite{chung16b}, LRS3~\cite{Afouras18d} and AVspeech~\cite{DBLP:journals/tog/EphratMLDWHFR18} as described in~\cite{ma2022visual}.

\begin{table}[t!]
\centering
\small
\begin{tabularx}{0.99\linewidth}{k g y k}
\toprule
\multirow{2}{*}[-0.2em]{\textbf{\shortstack{Self-Distillation \\ Models}}} & \multicolumn{3}{c}{\textbf{Top-1 Acc.} (\%)	} \\ 
& Scratch & LiRA(LRS3) & LRS2\&3+AVS \\
\cmidrule(l{4pt}r{4pt}){1-1}
\cmidrule(l{4pt}r{4pt}){2-4} 
Teacher & 92.1 &92.3 &92.9 \\
Student 1 & 92.5 &92.8 &93.5 \\
Student 2 & 92.8 &92.9 &93.5 \\
Student 3 & 92.5 &93.0 &93.5 \\
Student 4 &	-    &92.9 &93.3 \\ 
\cmidrule(l{4pt}r{4pt}){1-1}
\cmidrule(l{4pt}r{4pt}){2-4} 
Ensemble & \textbf{93.4} &\textbf{93.6} & \textbf{94.1} \\
\bottomrule
\end{tabularx}
\caption{Performance of self-distillation models (Teacher = ResNet-18 + DC-TCN). The best-performing models from Table \ref{tab:distentangling experiments on lrw} are serving as  teachers in first row.
For each student model, the model from the line above is used as its teacher, and ``Student $i$'' stands for the model after the $i$-th self-distillation iteration.}
\label{tab:self-distill}
\vspace{-0.8cm}
\end{table}
\begin{table}[t!]
\centering
\small
\begin{tabularx}{0.99\linewidth}{l y y}
\toprule
\textbf{Method} & \textbf{Word Boundary} & \textbf{Top-1 Acc.} (\%) \\ \midrule
3D-CNN~\cite{chung16} & \multirow{12}{*}[-0.4em]{
\xmark
}	& 61.1 \\ 
ResNet-34 + BLSTM~\cite{stafylakis2017combining} & & 83.0\\ 
2*3D-CNN + BLSTM~\cite{weng19} & & 84.1\\ 
ResNet-18 + BLSTM~\cite{stafylakis18} & & 84.3\\ 
ResNet-18 + BGRU + Cutout~\cite{zhang2020can} & & 85.0\\
ResNet-18 + BGRU~\cite{DBLP:journals/corr/abs-2011-07557} & & 85.0 \\ 
ResNet-18 + MS-TCN \cite{martinez2020lipreading} & & 85.3 \\ 
ResNet-18 + MS-TCN + S.D. \cite{ma2020towards} & & 88.5 \\  
ResNet-18 + DC-TCN \cite{ma2021lip} & & 88.4 \\ 
\cmidrule(lr){1-1} \cmidrule(lr){3-3}
Ours (w/o S.D., Scratch) & & \textbf{90.4} \\
Ours (w/o S.D., LRS2\&3+AVS) & & \textbf{91.1} \\
\cmidrule(lr){1-1} \cmidrule(lr){3-3}
Ours (Ensemble, Scratch) & & \textbf{91.6} \\
Ours (Ensemble, LRS2\&3+AVS) & & \textbf{92.1} \\ \midrule \midrule
ResNet-18 + BGRU~\cite{DBLP:journals/corr/abs-2011-07557} & \multirow{8}{*}[-0.4em]{\cmark} & 88.4 \\ 
ResNet-18 + BLSTM~\cite{stafylakis18} & & 88.8 \\ 
\cmidrule(lr){1-1} \cmidrule(lr){3-3}
Ours (w/o S.D., Scratch) & & \textbf{92.1} \\
Ours (w/o S.D., LiRA(LRS3)) & & \textbf{92.3} \\
Ours (w/o S.D., LRS2\&3+AVS) & & \textbf{92.9} \\
\cmidrule(lr){1-1} \cmidrule(lr){3-3}
Ours (Ensemble, Scratch) & & \textbf{93.4} \\
Ours (Ensemble, LiRA(LRS3)) & & \textbf{93.6} \\
Ours (Ensemble, LRS2\&3+AVS) & & \textbf{94.1} \\ 
\bottomrule
\end{tabularx}
\caption{Comparison with state-of-the-art methods on the LRW dataset in terms of classification accuracy. Experiments are divided into two groups, with and without utilising word boundaries indicators, respectively. ``S.D.'': self-distillation. 
\protect\say{Scratch}, \protect\say{LiRA(LRS3)} and \protect\say{LRS2\&3+AVS} correspond to the three pre-training strategies in Table \ref{tab:distentangling experiments on lrw}.}

\label{tab:sota_visual_lrw}
\vspace{-0.6cm}
\end{table}
\section{Results}

\noindent\textbf{Ablation Study}\quad
Results for the ablation study are shown in Table~\ref{tab:distentangling experiments on lrw}.
By removing one augmentation at a time we can estimate its contribution to the final model. We see the Time Masking is the most important augmentation, resulting in an absolute drop of 2.4\% followed by mixup with a drop of 1.1\%.
By replacing DC-TCN with MS-TCN, we observe that the performance drops by 2.1\,\%, which demonstrates the importance of dense connections and the SE attention mechanism in DC-TCN. The performance  drops by 2.4\% by replacing DC-TCN with BGRU. Additionally, the removal of word boundary indicators drops the performance by 1.7\,\%, which demonstrates the benefits of including auxiliary boundary indicators. Finally, we pre-train the encoder in a self-supervised/supervised manner on the LRS3\,/\,LRS2, LRS3 and AVspeech datasets and then fine-tune the  model on the LRW training set, and this slightly increases the performance to 92.3\,\%\,/\,92.9\,\%. It is clear from Table \ref{tab:sota_visual_lrw} that the proposed models significantly outperform the current state-of-the-art.

\noindent\textbf{Self-Distillation}\quad
Results for self-distillation experiments are presented in Table~\ref{tab:self-distill}. We use the best two models from Table \ref{tab:distentangling experiments on lrw} as teachers in the first round. It is clear that self-distillation results in a 0.6 \,\% to 0.7\,\%  absolute improvement in all cases. In addition, an ensemble of all models (all students + teacher) leads to a further absolute improvement of 0.6\,\%. These results suggest that self-distillation is beneficial for lip-reading. However, we should point out that the improvement is smaller compared to~\cite{ma2020towards}, probably due to the much better teacher model which makes further improvement harder.

\begin{figure}[!t]
  \centering
  \includegraphics[width=0.97\linewidth]{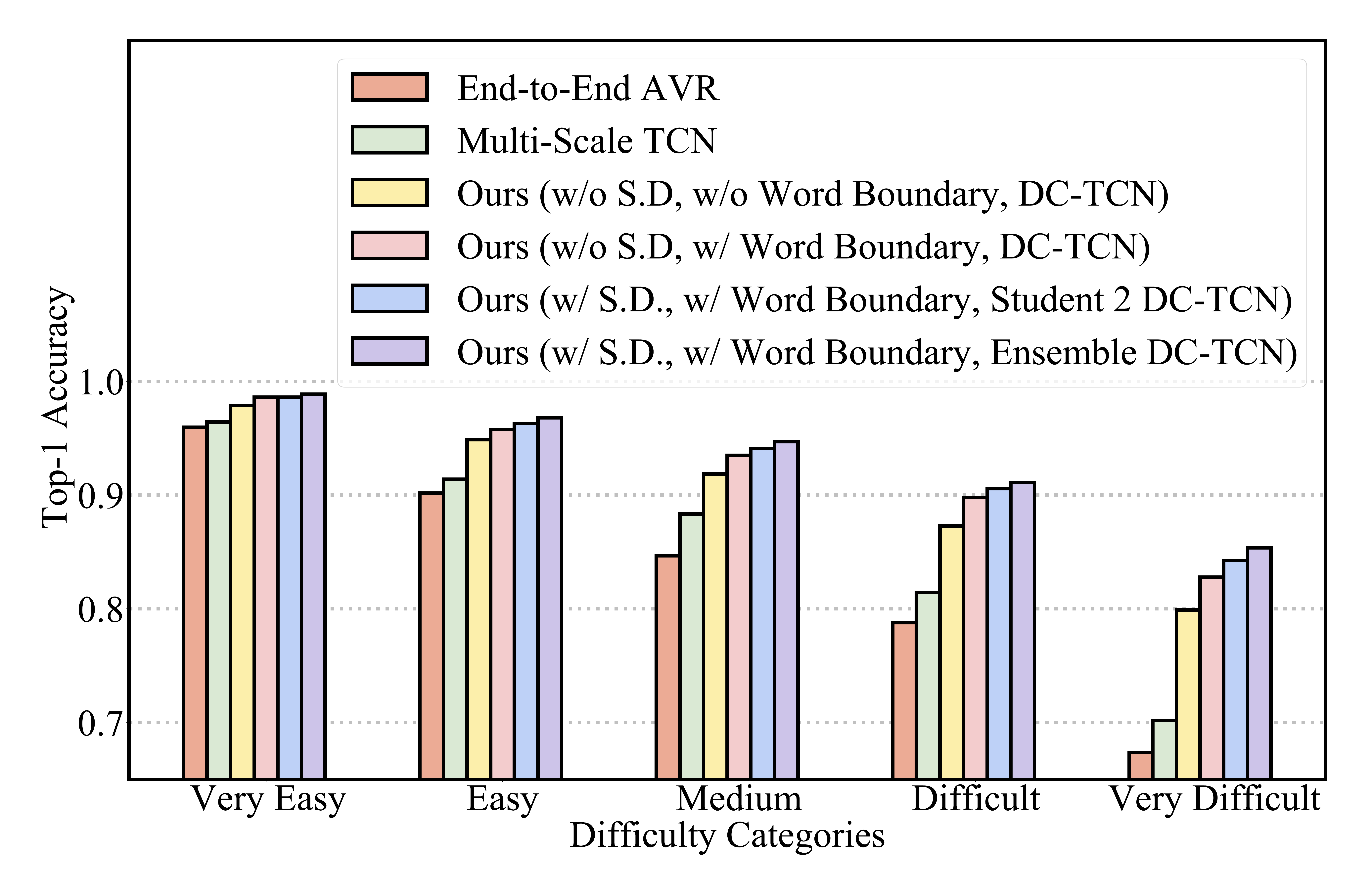}
  \vspace{-0.5cm}
  \caption{A comparison of our method and two baseline methods (End-to-End AVR~\cite{petridis2018end} and Multi-Scale TCN~\cite{martinez2020lipreading}) on the five difficulty groups of the LRW test set. }
  \label{difficulty_comparsion}
  \vspace{-0.4cm}
\end{figure}

\noindent\textbf{Error Analysis}\quad
In order to better understand how the presented models improve the word classification accuracy, we perform some error analysis. We divide the test samples in the LRW dataset into five groups~\cite{ma2021lip}. Each group contains 100 distinct isolated words and it is created based on the word accuracy of  the model in ~\cite{petridis2018end}. The 100 words with the highest classification accuracy are grouped in the \say{Very Easy} group, the next 100 words in the \say{Easy} group and so on. The average classification accuracy in each group is shown in Fig. \ref{difficulty_comparsion}. For comparison purposes, we also include the performance of ~\cite{martinez2020lipreading} and~\cite{petridis2018end}. We can see that our models outperform the two baselines across all groups and the  improvement is more pronounced in the \say{Difficult} and \say{Very Difficult} groups. 

\section{Conclusion}
In this work, we present a detailed study on the LRW dataset in terms of data augmentation and temporal models and demonstrate how state-of-the-art performance can be achieved by combining the best augmentations and training strategies. We show that Time Masking is the most important data augmentation method followed by mixup. We also show that DC-TCNs result in better performance than MS-TCNs or BGRUs. The use of self-distillation and word boundary indicators further improves the classification accuracy whereas the use of pre-training leads to a slight improvement. Finally, an error analysis reveals that the presented models significantly improve the classification accuracy of hard-to-recognise words.

\clearpage
\section{References}
\begingroup
\printbibliography[heading=none]
\endgroup

\end{document}